\documentclass[conference]{IEEEtran}
\usepackage{times}

\usepackage[numbers]{natbib}
\usepackage{multicol}
\usepackage[bookmarks=true]{hyperref}
\usepackage{xcolor}
\usepackage{capt-of}

\usepackage{multirow}
\usepackage{graphicx}
\usepackage{amsmath}
\usepackage{amssymb}
\usepackage{mathtools}
\usepackage{amsthm}

\begin{document}

\title{Self-Supervised Unseen Object Instance Segmentation via Long-Term Robot Interaction}

% You will get a Paper-ID when submitting a pdf file to the conference system
% \author{Author Names Omitted for Anonymous Review. Paper-ID [22]}

\author{Yangxiao Lu$^1$ \hspace{3px} Ninad Khargonkar$^1$ \hspace{3px}  Zesheng Xu$^1$ \hspace{3px} Charles Averill$^1$ \hspace{3px} Kamalesh Palanisamy$^1$ \\
Kaiyu Hang$^2$ \hspace{3px} Yunhui Guo$^1$ \hspace{3px} Nicholas Ruozzi$^1$ \hspace{3px} Yu Xiang$^{1}$\\
$^1$The University of Texas at Dallas \hspace{6px} $^2$Rice University\\
\tt\small $^1$\{firstname.lastname\}@utdallas.edu \tt\small $^2$\{firstname.lastname\}@rice.edu
}

%\author{\authorblockN{Michael Shell}
%\authorblockA{School of Electrical and\\Computer Engineering\\
%Georgia Institute of Technology\\
%Atlanta, Georgia 30332--0250\\
%Email: mshell@ece.gatech.edu}
%\and
%\authorblockA{Twentieth Century Fox\\
%\and
%\authorblockN{James Kirk\\ and Montgomery Scott}
%\authorblockA{Starfleet Academy\\
%San Francisco, California 96678-2391\\

% avoiding spaces at the end of the author lines is not a problem with
% conference papers because we don't use \thanks or \IEEEmembership

%Homer Simpson\authorrefmark{2},
%Montgomery Scott\authorrefmark{3} and
%Eldon Tyrell\authorrefmark{4}}
%Telephone: (800) 555--1212, Fax: (888) 555--1212}
%\authorblockA{\authorrefmark{4}Tyrell Inc., 123 Replicant Street, Los Angeles, California 90210--4321}}

\maketitle

\begin{abstract}

We introduce a novel robotic system for improving unseen object instance segmentation in the real world by leveraging long-term robot interaction with objects. Previous approaches either grasp or push an object and then obtain the segmentation mask of the grasped or pushed object after one action. Instead, our system defers the decision on segmenting objects after a sequence of robot pushing actions. By applying multi-object tracking and video object segmentation on the images collected via robot pushing, our system can generate segmentation masks of all the objects in these images in a self-supervised way. These include images where objects are very close to each other, and segmentation errors usually occur on these images for existing object segmentation networks. We demonstrate the usefulness of our system by fine-tuning segmentation networks trained on synthetic data with real-world data collected by our system. We show that, after fine-tuning, the segmentation accuracy of the networks is significantly improved both in the same domain and across different domains. In addition, we verify that the fine-tuned networks improve top-down robotic grasping of unseen objects in the real world \footnote{Video, dataset and code are available at \url{https://irvlutd.github.io/SelfSupervisedSegmentation}}.

\end{abstract}

\IEEEpeerreviewmaketitle

\section{Introduction}

Object perception is a critical task in robot manipulation. Model-based methods leverage 3D models of objects and solve the 6D object pose estimation problem to localize objects in 3D~\cite{collet2011moped,xiang2017posecnn,tremblay2018deep,wada2020morefusion}. Using the estimated object poses and the 3D models of objects, a planning scene can be set up for manipulation trajectory planning. However, requiring a 3D model for every object that needs to be manipulated is not feasible in the real world. Recent model-free approaches for object perception focus on segmenting unseen objects from images~\cite{xiang2021learning,xie2021unseen,durner2021unknown}. A segmented point cloud of an object can be used in grasp planning for robot manipulation~\cite{mousavian20196,sundermeyer2021contact}. In this way, an object can be grasped from partial observations without using its 3D model.

Recent model-based and model-free methods for object perception train neural networks to recognize objects. Since it is difficult to obtain large-scale real-world datasets in robot manipulation settings, synthetic data is widely used for training~\cite{tremblay2018training,xie2020best,back2022unseen}. Although models trained with synthetic data can be directly used in the real world by leveraging domain randomization~\cite{tobin2017domain} or domain transfer~\cite{balloch2018unbiasing,zhang2022unseen} techniques, these models still have errors in the real world due to the sim-to-real gap. The question we would like to address in this paper is how can a robot automatically obtain training data in the real world to improve its object segmentation model pre-trained with synthetic data. We focus on improving Unseen Object Instance Segmentation (UOIS) to facilitate robot manipulation.

Interactive perception~\cite{bohg2017interactive} emphasizes that robots can apply actions to the environments and utilize the visual-motor relationship to improve perception. In the context of object recognition, two widely used interaction types are robot grasping and pushing. Previous works have explored leveraging robot grasping or pushing to obtain object segmentation data in a self-supervised way~\cite{pathak2018learning,eitel2019self,yu2022self}. All these methods can only obtain the segmentation mask of the grasped or pushed object by comparing the scene before and after grasping~\cite{pathak2018learning} or utilizing optical flow to segment the moved objects in robot pushing~\cite{eitel2019self,yu2022self}. The drawbacks of segmenting objects from one action are that, first, the method cannot segment unmoved objects in the scene; second, if two objects are moved together, the method will segment them as one object. Although~\cite{yu2022self} proposes to train a classifier to decide whether a single object is pushed or not, since the classifier is trained in simulation, it still suffers from the sim-to-real gap.

\begin{figure}
\begin{center}
\includegraphics[width=0.9\linewidth]{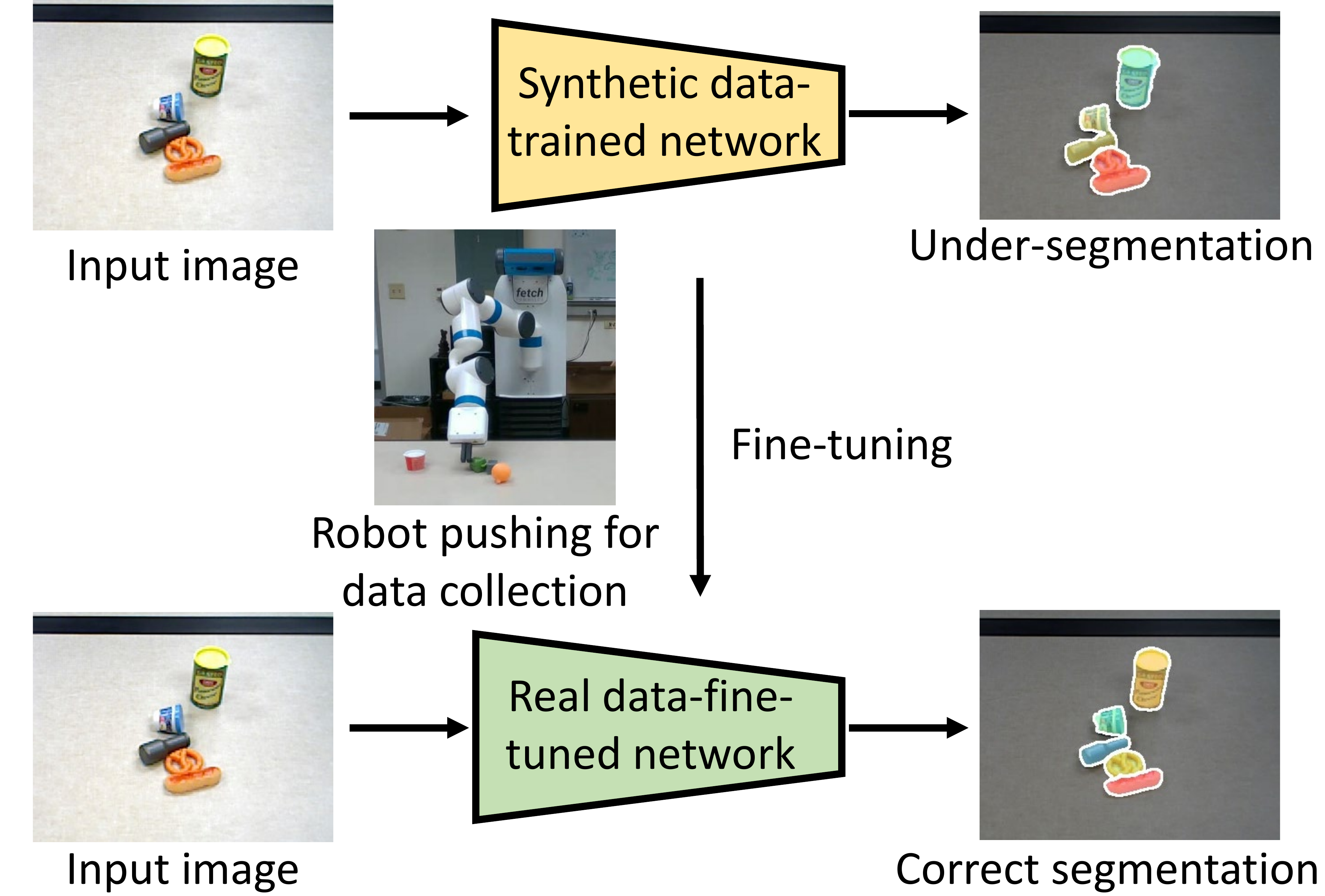}
\caption{Our system leverages robot pushing to collect real-world images and generate segmentation masks of objects in the collected images in a self-supervised way. The collected images can be used to fine-tune segmentation networks trained with synthetic data and improve their performance.}
\label{fig:intro}
\end{center}
\vspace{-5mm}
\end{figure}

To overcome the limitations of existing work on self-supervised object segmentation via robot interaction, we propose a new system that leverages long-term robot interaction to segment unseen objects in a self-supervised way. Our key idea is to defer the decision on object segmentation until a robot has interacted with all the objects in a scene for a period of time. Intuitively, if a robot has pushed objects in a scene for a number of times, i.e., around 20 pushes for 5 objects in our experiments, these objects are very likely to be separated from each other. Once the objects are separated, existing approaches on unseen object segmentation such as~\cite{xiang2021learning,lu2022mean} can successfully segment them. In this way, our system can segment all the objects in the scene but not only the pushed object in one action. More importantly, the system enables the robot to propagate a correctly segmented mask of each object to all the collected images during robot pushing including images where objects are very close to each other. This is achieved by combining multi-object tracking to extract object tracklets, i.e., segments of objects in video frames, and video object segmentation where an initial mask of an object can be propagated to all other frames. The system utilizes the object tracklet to select a good initial mask for propagation. Consequently, our system enables a robot to collect a sequence of images of objects in a scene and obtain segmentation masks of all the objects in these images.

We demonstrate the usefulness of our system by using the collected real-world images to fine-tune existing, pre-trained object segmentation models~\cite{lu2022mean}. We show that after fine-tuning, the object segmentation accuracy of the model can be significantly improved. The improvement is achieved in the same domain as the fine-tuning data as well as on the benchmark datasets for evaluating unseen object instance segmentation~\cite{richtsfeld2012segmentation,suchi2019easylabel}. Fig.~\ref{fig:intro} illustrates the fine-tuning process. In addition, we show that using the fine-tuned segmentation model can improve top-down grasping performance in a table clearing task where a robot is asked to put all the objects on a table into a bin.
In summary, the contributions of our work are as follows.
\begin{itemize}
    \item We introduce a novel robotic system that leverages long-term robot interaction to segment unseen objects in a self-supervised way.

    \item Our system illustrates that combining multi-object tracking and video object segmentation with robot pushing can help robots to singulate objects from each other in cluttered scenes.

    \item We demonstrate that using our system to collect real-world images for fine-tuning can improve object segmentation accuracy and robot grasping performance.
    
\end{itemize}

\section{Related Work}

\subsection{Unseen Object Instance Segmentation}

Different from category-based object instance segmentation methods~\cite{he2017mask,chen2017deeplab,cheng2022masked} that focus on segmenting object instances among a set of pre-defined object categories, unseen object instance segmentation emphasizes segmenting arbitrary objects that present in input images. The testing objects can be novel such that a segmentation model has not seen them during training. Earlier works on UOIS utilize low-level image cues such as edges, contours, and surface normals to group pixels into objects~\cite{richtsfeld2012segmentation,trevor2013efficient,christoph2014object}. These bottom-up approaches tend to over-segment objects since there is no object-level supervision to learn the concept of objects. Recent approaches on UOIS leverage large-scale synthetic data and deep neural networks to segment unseen objects~\cite{xie2020best,xiang2021learning,xie2021unseen,durner2021unknown}. These methods significantly improve object segmentation accuracy, which enables robotic grasping of unseen objects~\cite{mousavian20196,sundermeyer2021contact}. However, since these models are trained with synthetic data, they still suffer from the sim-to-real gap. The primary error is under-segmentation in the real world. When objects are very close to each other, the models trained with synthetic data cannot separate them. Recently, \citet{zhang2022unseen} propose to apply test-time domain adaption to improve the segmentation performance, where a set of images without ground truth labels in the test domain are used to adapt the segmentation network. Our system is complementary to domain adaption techniques since it is able to obtain training images with ground truth labels automatically. Therefore, we can use supervised learning to fine-tune segmentation networks. More importantly, we show that, after fine-tuning in one domain, the performance of the segmentation networks can be improved in other domains, which avoids adaption in every testing domain. 

\subsection{Self-Supervised Robot Perception}

Self-supervised learning is an attractive learning paradigm where training data and training signals can be obtained automatically without human labor. Since a robot can naturally interact with its environment to collect data~\cite{bohg2017interactive}, self-supervised learning for robot perception has received more attentions recently. One type of approach utilizes multi-view consistency of images captured from different viewpoints to obtain the ground truth annotations for learning. Multi-view consistency based self-supervised learning has been applied to object segmentation~\cite{zeng2017multi}, object detection~\cite{mitash2017self}, 6D object pose estimation~\cite{deng2020self} and dense pixel-wise correspondences~\cite{schmidt2016self,florence2018dense} in robot manipulation settings. Another type of approach leverages robot actions such as grasping and pushing to interact with objects and then computes scene differences~\cite{pathak2018learning} or optical flow~\cite{eitel2019self,yu2022self} before and after applying an action to obtain ground truth labels of objects for learning. Our system falls into this category where we also employ robot pushing with optical flow to help segment objects in a self-supervised way. The main novelty of our system compared to previous methods on self-supervised object segmentation~\cite{eitel2019self,yu2022self} is that we leverage long-term robot pushing to segment all the objects in a collected video sequence, while previous methods can only segment the grasped or pushed object in an image. 

\section{Self-Supervised Unseen Object Instance Segmentation}

\subsection{System Overview}

\begin{figure*}
\begin{center}
\includegraphics[width=0.95\linewidth]{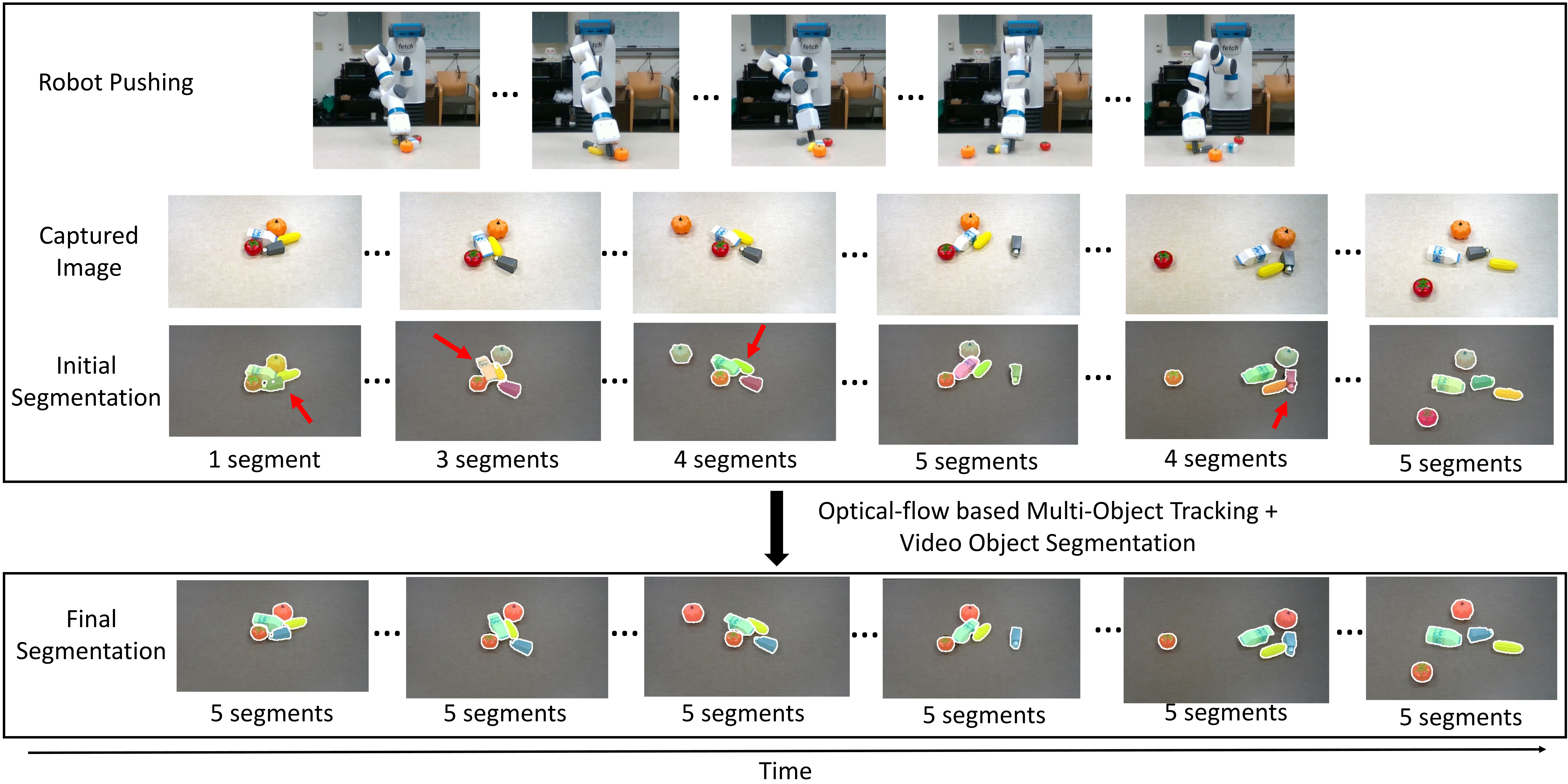}
\caption{System overview. Our system leverages robot pushing to interact with objects. The pushing actions are guided by the initial segmentation masks of the objects generated from a segmentation network trained with synthetic data. Images before and after each pushing action are captured. By using the sequence of images with the initial segmentation masks, our system combines optical flow-based multi-object tracking and video object segmentation to compute the final segmentation masks, which fix errors in the initial segmentation masks. Red arrows indicate the segmentation errors. The collected images and the final segmentation masks can be used to fine-tune the segmentation network to improve its performance.}
\vspace{-4mm}
\label{fig:overview}
\end{center}
\end{figure*}

The motivation to build our system is to fix segmentation errors in existing UOIS methods~\cite{xiang2021learning,lu2022mean}. These methods are trained with synthetic RGB-D images generated using 3D models of objects. Due to the sim-to-real gap and the arrangements of objects in the simulator, these methods often cannot separate objects that are very close to each other. One example is shown in the first initial segmentation image in Fig.~\ref{fig:overview}, where five objects are packed together and the MSMFormer \cite{lu2022mean} only outputs one mask for all five objects. In grasping applications, a robot cannot grasp these objects due to the incorrect segmentation result. Our idea to fix these errors is to obtain ground truth masks of these packed objects in a self-supervised way by leveraging robot interaction with objects. Then, we can use these images with the corresponding ground truth masks to fine-tune the segmentation networks~\cite{xiang2021learning,lu2022mean}. With enough data for fine-tuning, the networks should be able to segment closely packed objects.

The main challenge in this scenario is obtaining the ground truth masks when objects are close to each other. Previous methods that leverage robot interaction to obtain object masks~\cite{eitel2019self,yu2022self} can only obtain one mask of the pushed or grasped object in an image. They cannot generate masks of all the objects in the scene because they only use one robot action and try to figure out which object has been moved. Instead, in our system, we allow the robot to continuously push objects in a random fashion, and we generate a sequence of images before and after each pushing action, i.e., around 20 pushes for each scene in our experiments. Finally, we use these images to perform multi-object tracking and video object segmentation. In this way, our system can generate masks of all the objects in the image sequence including the first image, where all the objects are close to each other. Fig.~\ref{fig:overview} illustrates an overview of our system. The collected images with their generated masks can be used to fine-tune existing methods for unseen object instance segmentation~\cite{xiang2021learning,lu2022mean} in order to improve their performance in the real world. We introduce each component of the system in the following sections.

\subsection{Data Collection via Robot Pushing}

Since our goal is to collect hard to segment images to fine-tune the segmentation networks, we intentionally put objects together for each scene in the beginning of the data collection process. After setting up a scene on a tabletop, a robot starts pushing these objects. A Fetch mobile manipulator is employed in our system, and an RGB-D image is captured before and after each push action, where we used the RGB-D camera on the Fetch robot to capture images.

Different from methods that carefully learn a pushing or grasping policy for singulation~\cite{yu2022self}, we design a simple pushing strategy using object instance segmentation from the MSMFormer~\cite{lu2022mean} as input. This is because our system does not require all the objects to be singulated at the end of the interaction. As long as an object has been separated from other objects for a period of time during pushing, the system is able to generate correct segmentation masks for it thanks to the multi-object tracking and video object segmentation techniques utilized in the system. In cases where one push action cannot separate two objects if both objects move together, multiple push actions may separate them. Therefore, our system benefits from long-term robot interactions with a sequence of pushes.

Specifically, suppose at time $t$, the system captures an RGB-D image $I_t$. We obtain a set of $n_t$ object segmentation masks $\{ o_t^i \}_{i=1}^{n_t}$ on $I_t$ by running the MSMFormer network on it. These masks are illustrated as the initial segmentation in Fig.~\ref{fig:overview}. Based on the object segmentation, the robot randomly selects an object to push. First, a 3D bounding box is computed for each segmented object by bounding the 3D point cloud of the object. Using the depth image and the camera intrinsic parameters, we can back-project the depth image into a 3D point cloud of the scene in the camera frame. Since we also know the camera pose in the robot frame, we can convert the point cloud into the robot frame. Using the segmentation mask of each object, we can extract the points of the object and compute a 3D bounding box for it in the robot frame. Second, according to center of the 3D bounding box, the robot decides to either push the object to the left or to the right. We select the pushing direction to always push the object towards the center of the robot, which prevents objects being pushed outside the reach of the robot. Third, a motion trajectory is planned to the left side (pushing right) or right side (pushing left) of the object. We used the MoveIt motion planning framework to plan the trajectories. Then the planned trajectory is executed to move the robot arm to the pushing location. Finally, the pushing action is achieved by adding an offset to the shoulder joint of the Fetch arm depending on the pushing direction.

Note that our pushing strategy cannot achieve perfect singulation results compared to learned polices or designed strategies for singulation. However, singulation is not our main goal. We also want to collect diverse datasets for learning. Our pushing strategy is effective to separate objects and perturb objects in the scene in order to generate diverse images. In addtion, although the initial segmentation has errors, it can still be used to guide the pushing process. A sequence of pushing actions and the generated images are shown in Fig.~\ref{fig:overview}.

\begin{figure}
\begin{center}
\includegraphics[width=0.9\linewidth]{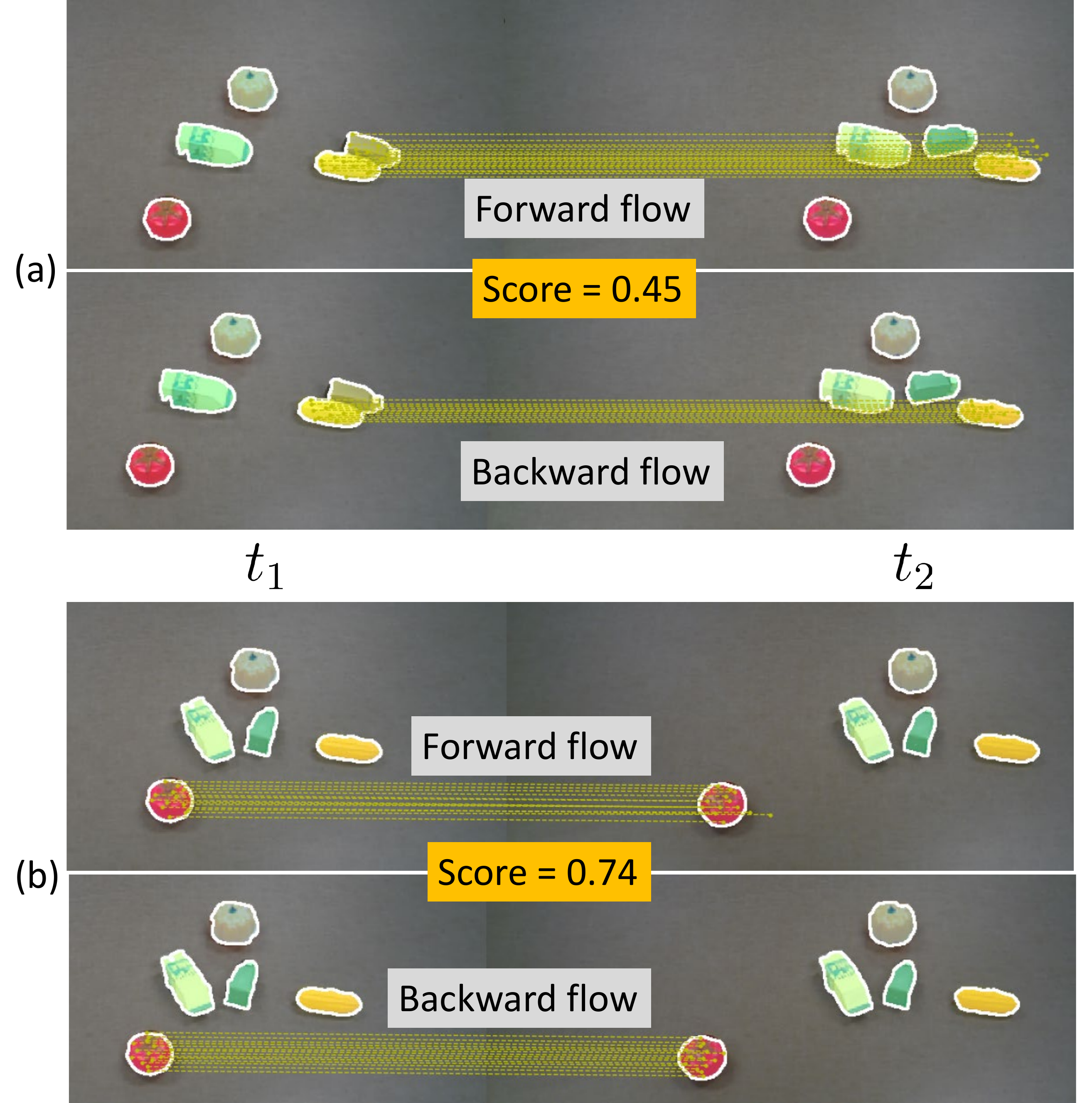}
\caption{Illustration of the matching scores between objects based on forward and backward optical flow.}
\label{fig:score}
\end{center}
\vspace{-6mm}
\end{figure}

\subsection{Optical Flow-based Multi-Object Tracking}

After the data collection via robot pushing, we obtain a sequence of images $I_1, I_2, \ldots, I_N$ with the corresponding initial segmented objects $\{ o_1^i \}_{i=1}^{n_1}, \{ o_2^i\}_{i=1}^{n_2}, \ldots, \{ o_N^i \}_{i=1}^{n_N}$, where $N \approx 20$ in our experiments. Since there are errors in these initial masks, our next task is to fix these errors and obtain correct segmentation masks for all the objects in the image sequence. Our idea is to leverage the observation that if a mask incorrectly includes more than one object, after a robot pushing, the mask will be broken down into multiple objects. On the other hand, if a mask correctly segments one object, after pushing, the mask will remain the same. However, one pushing action may not be able to singulate an object successfully. Therefore, we leverage a sequence of robot pushing actions in our system. In this case, if a mask remains the same after several pushing actions, it is highly likely to be a correct segmentation.

In order to compare the initial segmentation masks across image frames, we need to associate masks across frames. This problem is studied in the literature as tracking by detection~\cite{zhang2008global,andriluka2008people,xiang2015learning,schulter2017deep}. The most important component in a tracking-by-detection method is a similarity measurement between two object detections across video frames, which can be learned from data~\cite{schulter2017deep} or defined using image features~\cite{xiang2015learning}. In our system, since we do not have many data to learn the similarity measurement in robotic manipulation settings, we design one based on optical flow between image frames.

Let $o_{t_1}^i$ be a mask on image $I_{t_1}$ and $o_{t_2}^j$ be a mask on image $I_{t_2}$. We would like to compute a similarity score between the two masks as $s(o_{t_1}^i, o_{t_2}^j)$. We only consider adjacent images in data association. Therefore, we can assume $t_2 = t_1 + 1$. We leverage optical flow between the two images to define the similarity score. Let $o_{t_2}^i = o_{t_1}^i + f_{t_1, t_2}^i$ be the propagated mask of object  $o_{t_1}^i$ to frame $I_{t_2}$ using forward flow $f_{t_1, t_2}^i$. Similarly, we can propagate the mask of object $o_{t_2}^j$ to frame $I_{t_1}$ using backward flow: $o_{t_1}^j = o_{t_2}^j + f_{t_2, t_1}^j$. The similarity score between the two masks is defined as
\begin{equation} \label{eq:score}
    s(o_{t_1}^i, o_{t_2}^j) = \min \big( \text{IoU}(o_{t_2}^i, o_{t_2}^j), \text{IoU}(o_{t_1}^i, o_{t_1}^j) \big),
\end{equation}
where the $\text{IoU}(\cdot, \cdot)$ function computes the intersection over union between two binary masks. Intuitively, one mask is propagated to another image using optical flow and compared to the other mask.

\begin{figure*}
\begin{center}
\includegraphics[width=0.9\linewidth]{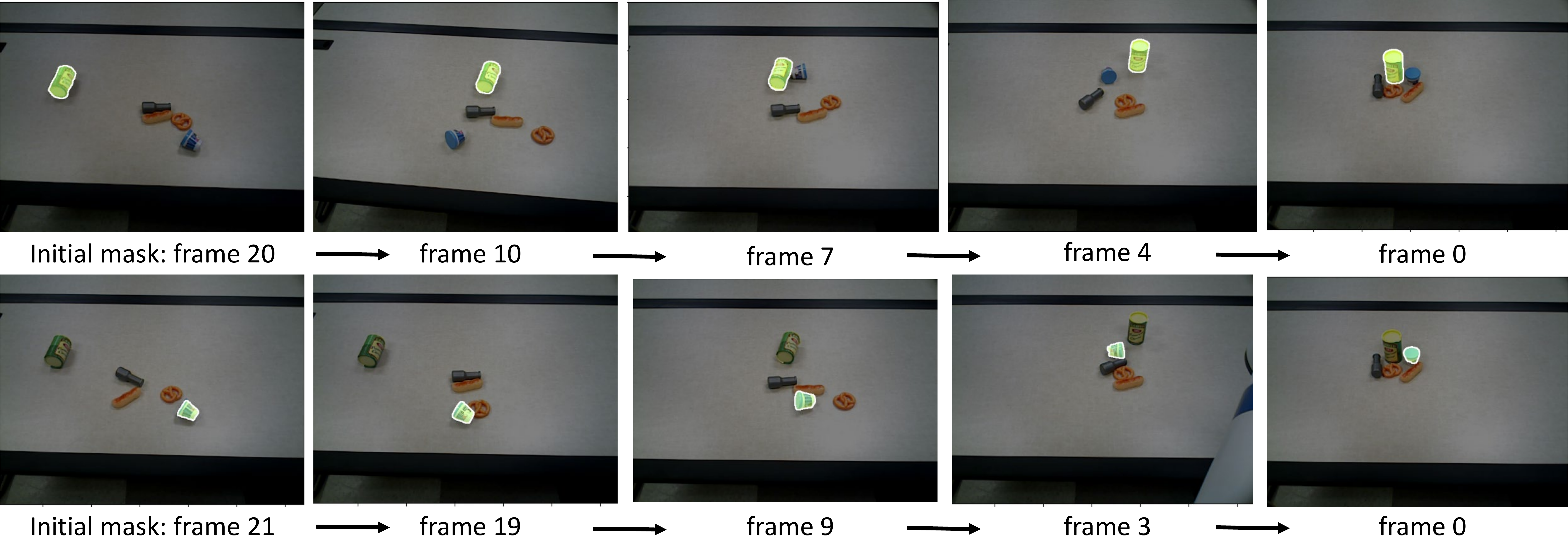}
\caption{Illustration of the XMem~\cite{cheng2022xmem} video object segmentation in our collected data. The initial mask is used to initialize the segmentation process.}
\label{fig:xmem}
\end{center}
\vspace{-6mm}
\end{figure*}

Fig.~\ref{fig:score} illustrates two examples of the computed matching scores. In case (a), at time $t_1$, the initial segmentation cannot separate the corn and the salt bottle. The propagated mask to time $t_2$ cannot match the mask of the corn at time $t_2$ well. Therefore, the matching score is low. In case (b), the masks of the tomato match well using both the forward flow and the backward flow. The matching score is high. When the optical flow estimation is accurate, the similarity score in Eq.~\eqref{eq:score} serves as a good measurement for data association between objects. In our system, we use the RAFT~\cite{teed2020raft} network to compute optical flow.

With the above similarity score, we can leverage existing multi-object tracking methods such as network flow-based approaches~\cite{zhang2008global,schulter2017deep} or Markov decision process-based approaches~\cite{xiang2015learning} to generate trajectories of objects across image frames. Instead, we found that a simple greedy search algorithm works well in the tabletop robot pushing settings since there are no long-term occlusions between objects or new objects coming in and out in these settings. The greedy data association algorithm starts from one mask in the last image frame $I_N$. Then it associates the mask to a previous mask which has the highest matching score if their matching score is larger than a pre-defined threshold, and repeats this process until the highest matching score is smaller than the threshold. In this way, it generates a tracklet for one object. After that, it selects a remaining mask and repeats the process to generate the next tracklet. We start the data association from the last frame in a backward way because objects are likely to be separated in the end of the robot pushing, which helps for object tracking.

\subsection{Mask Propagation via Long-Term Object Segmentation}

The output from the multi-object tracking algorithm is a set of tracklets $\{ \mathcal{T}_i \}_{i=1}^M$, where tracklet $\mathcal{T}_i = (o_{t_1}^i, o_{t_2}^i, \ldots, o_{t_m}^i )$ consists of a sequence of object masks from the initial segmentation. The lengths of these tracklets can be different. Majority masks in each tracklet correctly segment one object, since wrong initial segmentation masks have low matching scores as illustrated in Fig.~\ref{fig:score}. If we can utilize the extracted tracklets and propagate the correct masks to all the image frames for all the objects, we can obtain correct segmentation masks for the collected data via robot pushing.

To achieve this goal, we utilize a state-of-the-art video object segmentation method named XMem~\cite{cheng2022xmem}. Given an initial mask of an object, XMem can segment the object in the following video frames. It maintains a memory buffer that stores the features of the target object, which enables it to segment the target in long video sequences and handle occlusions. In the traditional video segmentation scenarios, the initial mask of a target is given manually on the first video frame. In our case, we need to generate the initial mask automatically. It is critical to select a correct initial mask for an object. Otherwise, a wrong mask will be propagated to other frames. We utilize the observation that if a mask being pushed can still have high matching scores (Eq.~\eqref{eq:score}) to the previous mask and the next mask in a tracklet, the mask is likely to contain a single object. Therefore, we select the pushed mask with the highest matching score as the initial mask to initialize XMem. The segmentation goes with two directions, where one goes to the first frame and the other one goes to the last frame in the collected image sequence. Fig.~\ref{fig:xmem} shows two examples of the object segmentation with XMem. After all the tracklets are processed, the segmentation masks are combined to generate the final segmentation of the images (see Fig.~\ref{fig:overview}). In this way, our system can obtain segmentation masks of objects when they are very close to each other.

\section{Applications}

\subsection{Transfer Learning for Object Segmentation}

Our system can be used to collect images with the corresponding object segmentation masks in a self-supervised way. Then we can use these images to fine-tune the object segmentation networks to improve their performance. Since the collected data include correct segmentation masks when objects are very close to each other, the fine-tuned model is able to fix segmentation errors and correctly separate objects in cluttered scenes.

For the fine-tuning, we start with a segmentation model trained with synthetic data. We used  MSMFormer~\cite{lu2022mean} in our experiments, which is also used to generate the initial segmentation masks for robot pushing. We initialize the network with the pre-trained weights on the synthetic data, and then train the network for a number of epochs on the collected real-world data with a smaller learning rate. We conducted an ablation study on different fine-tuning strategies. Specifically, the backbone of the network can be fixed or be trainable during fine-tuning. The fine-tuning data can be a mixture of synthetic images and real-world images or real-world images only. The effect of these strategies are presented in Section~\ref{sec:exp}.

\begin{table*}[!ht]
\caption{UOIS first-stage RGB-D results on OCID and OSD with fixed or learnable backbone and different datasets. These strategies are used to fine-tune MSMFormer \cite{lu2022mean}.}\label{backbones}
\centering
\scalebox{1.0}{
\begin{tabular}{|c|lllllll|lllllll|}
\hline
\multirow{3}{*}{Datasets and backbones} & \multicolumn{7}{c|}{OCID (2390 images)}                                                                                                                                                 & \multicolumn{7}{c|}{OSD (111 images)}                                                                                                                                                   \\ \cline{2-15} 
                                        & \multicolumn{3}{c|}{Overlap}                                                       & \multicolumn{3}{c|}{Boundary}                                                      &               & \multicolumn{3}{c|}{Overlap}                                                       & \multicolumn{3}{c|}{Boundary}                                                      &               \\
                                        & \multicolumn{1}{c}{P} & \multicolumn{1}{c}{R} & \multicolumn{1}{c|}{F}             & \multicolumn{1}{c}{P} & \multicolumn{1}{c}{R} & \multicolumn{1}{c|}{F}             & \%75          & \multicolumn{1}{c}{P} & \multicolumn{1}{c}{R} & \multicolumn{1}{c|}{F}             & \multicolumn{1}{c}{P} & \multicolumn{1}{c}{R} & \multicolumn{1}{c|}{F}             & \%75          \\ \hline
MSMFormer \cite{lu2022mean}                                          & 88.4                  & \textbf{90.2}                  & \multicolumn{1}{l|}{88.5}          & 84.7                  & 83.1                  & \multicolumn{1}{l|}{83.0}          & 80.3          & 79.5                  & \textbf{86.4}                  & \multicolumn{1}{l|}{82.8}          & 53.5                  & 71.0                  & \multicolumn{1}{l|}{60.6}                               & 79.4          \\
Pushing Data + fixed backbone           & \textbf{93.9}         & 48.5                  & \multicolumn{1}{l|}{51.0}          & 80.7                  & 45.8                  & \multicolumn{1}{l|}{43.4}          & 47.9          & 81.4                  & 72.8                  & \multicolumn{1}{l|}{75.5}          & 41.1                  & 61.6                  & \multicolumn{1}{l|}{47.5}          & 65.3          \\
Pushing Data + learnable backbone       & 88.8                  & 82.6                  & \multicolumn{1}{l|}{85.3}          & 65.3                  & 72.4                  & \multicolumn{1}{l|}{68.2}          & 79.8          & \textbf{88.8}         & 82.6                  & \multicolumn{1}{l|}{\textbf{85.3}} & 65.3                  & \textbf{72.3}         & \multicolumn{1}{l|}{68.2}          & \textbf{79.8} \\
Mixture Data + fixed backbone           & 90.8                  & 88.2                  & \multicolumn{1}{l|}{88.9}          & 82.3                  & 82.9                  & \multicolumn{1}{l|}{81.9}          & 83.3          & 82.0                  & 85.0         & \multicolumn{1}{l|}{83.4}          & 53.9                  & 69.7                  & \multicolumn{1}{l|}{59.9}          & 78.4          \\

Mixture Data + learnable backbone       & 91.2                  & 90.1         & \multicolumn{1}{l|}{\textbf{90.1}} & \textbf{87.2}         & \textbf{85.5}         & \multicolumn{1}{l|}{\textbf{85.7}} & \textbf{83.9} & 85.1                  & 84.4                  & \multicolumn{1}{l|}{84.6}          & \textbf{67.8}         & 71.4                  & \multicolumn{1}{l|}{\textbf{69.0}} & 76.2          \\ \hline
\end{tabular}
}
\vspace{-2mm}
\end{table*}

\subsection{Top-Down Robot Grasping}
\label{sec:applications-top-down-grasping}

Unseen object instance segmentation can facilitate robot grasping of unknown objects as demonstrated in previous works~\cite{mousavian20196,murali20206}. These methods use the segmented point clouds of objects to plan grasps for grasping. Improvement in object segmentation can benefit the grasp planning stage and improve the grasping performance subsequently. In this work, we show that using our collected data for fine-tuning can improve object segmentation and top-down grasping consequently.

With accurate object segmentation, top-down grasp planning can be achieved in an analytic way. A top-down grasp for a two-finger gripper is defined as the 3D location $p = (x, y, z)$, orientation $\theta$ of the gripper in the $x,y$ plane and the width $w$ between the two fingers, where axis-$z$ is the gravity direction. The grasping position $p$ is defined as the object center, where the object center is computed as the mean of the segmented point cloud of the 
object. The grasping orientation $\theta$ is computed to align the gripper with the second largest principal component of the object point cloud in the $x,y$ plane. In this way, the robot can grasp the narrower side of a long object. Finally, the width between the two fingers is determined by the width of the object along the second largest principal component of the object point cloud in the $x,y$ plane. It can be shown that if the center of mass of the object is the same as the object center, a grasp computed in this way can achieve force closure. The described grasp planning algorithm relies on accurate segmentation of all the objects in a scene. We can use it to verify the benefit of our system in collecting data to improve object segmentation for robot grasping.

% Fig.~\ref{fig:grasp} illustrates three different cases in top-down grasp planning. When a target object is isolated from other objects, i.e., case (a) in Fig.~\ref{fig:grasp},  When the target object is close to some other objects, i.e., case (b) and case (c) in Fig.~\ref{fig:grasp}, the grasp will be shifted along the second largest principal component of the object in the $x,y$ plane such that one finger aligns with the boundary of the target and the other finger is in a collision-free position. In cases when the target object is surrounded by other objects on both sides, the robot can choose another object to grasp first.

\begin{table*}[!ht]
\caption{UOIS RGB-D results on OCID and OSD with different number of scenes for fine-tuning the first-stage MSMFormer model.}\label{scenes}
\centering
\scalebox{1.0}{
\begin{tabular}{|c|c|lllllll|lllllll|}
\hline
\multirow{3}{*}{\# of scenes} & \multirow{3}{*}{\begin{tabular}[c]{@{}l@{}}\# of images\end{tabular}} & \multicolumn{7}{c|}{OCID (2390 images)}                                                                                                                                                 & \multicolumn{7}{c|}{OSD (111 images)}                                                                                                                                                   \\ \cline{3-16} 
                              &                                                                                         & \multicolumn{3}{c|}{Overlap}                                                       & \multicolumn{3}{c|}{Boundary}                                                      &               & \multicolumn{3}{c|}{Overlap}                                                       & \multicolumn{3}{c|}{Boundary}                                                      &               \\
                              &                                                                                         & \multicolumn{1}{c}{P} & \multicolumn{1}{c}{R} & \multicolumn{1}{c|}{F}             & \multicolumn{1}{c}{P} & \multicolumn{1}{c}{R} & \multicolumn{1}{c|}{F}             & \%75          & \multicolumn{1}{c}{P} & \multicolumn{1}{c}{R} & \multicolumn{1}{c|}{F}             & \multicolumn{1}{c}{P} & \multicolumn{1}{c}{R} & \multicolumn{1}{c|}{F}             & \%75          \\ \hline
MSMFormer \cite{lu2022mean}                    & 0                                                                                       & 88.4                  & \textbf{90.2}         & \multicolumn{1}{l|}{88.5}          & 84.7                  & 83.1                  & \multicolumn{1}{l|}{83.0}          & 80.3          & 79.5                  & 86.4                  & \multicolumn{1}{l|}{82.8}          & 53.5                  & 71.0                  & \multicolumn{1}{l|}{60.6}          & 79.4          \\
3                             & 62                                                                                      & 89.7                  & 89.8                  & \multicolumn{1}{l|}{88.7}          & 82.8                  & 85.5                  & \multicolumn{1}{l|}{83.0}          & 85.3          & 83.6                  & 85.8                  & \multicolumn{1}{l|}{84.6}          & 58.7                  & 75.4                  & \multicolumn{1}{l|}{65.5}          & 80.6          \\
6                             & 124                                                                                     & 91.0                  & 89.1                  & \multicolumn{1}{l|}{89.5}          & 80.7                  & 85.0                  & \multicolumn{1}{l|}{82.0}          & \textbf{87.0} & 83.7                  & 85.1                  & \multicolumn{1}{l|}{84.3}          & 59.1                  & 74.6                  & \multicolumn{1}{l|}{65.3}          & 78.0          \\
9                             & 190                                                                                     & 91.4                  & 89.6                  & \multicolumn{1}{l|}{90.0}          & 83.7                  & \textbf{85.6}         & \multicolumn{1}{l|}{84.0}          & 86.0          & 83.9                  & 86.4                  & \multicolumn{1}{l|}{85.1}          & 58.6                  & 76.4                  & \multicolumn{1}{l|}{65.8}          & 81.0          \\
12                            & 256                                                                                     & \textbf{92.1}         & 89.7                  & \multicolumn{1}{l|}{\textbf{90.3}} & 86.2                  & 84.9                  & \multicolumn{1}{l|}{84.9}          & 86.3          & \textbf{87.6}         & \textbf{86.6}         & \multicolumn{1}{l|}{\textbf{87.0}} & 64.6                  & \textbf{77.5}         & \multicolumn{1}{l|}{\textbf{69.7}} & \textbf{85.6} \\
15 (All)                      & 321                                                                                     & 91.2                  & 90.1                  & \multicolumn{1}{l|}{90.1}          & \textbf{87.2}         & 85.5                  & \multicolumn{1}{l|}{\textbf{85.7}} & 83.9          & 85.1                  & 84.4                  & \multicolumn{1}{l|}{84.6}          & \textbf{67.8}         & 71.4                  & \multicolumn{1}{l|}{69.0}          & 76.2          \\ \hline
\end{tabular}
}
\end{table*}

\section{Experiments} \label{sec:exp}

\subsection{Datasets and Evaluation Metrics}

\textbf{Data Collected by the Robot.}
We used a set of play food for kids as the objects for robot interaction. For reproducibility, these objects can be purchased from~\cite{playfood}. A Fetch mobile manipulator is used for data collection. Five different objects are used in each scene, and the robot performs around 20 pushing actions for each scene to collect images before and after each pushing action. In total, we collected images from 20 scenes. Images from 15 scenes are used for fine-tuning and the remaining images are used for testing the fine-tuned model in the same domain. Specifically, 321 images are used for fine-tuning, while 107 images are available for testing. Each image contains an average of 6 objects, but no more than 8 objects.

% \yu{Count the nubmer of images for training and testing.}\lu{321 training and 107 testing}

\textbf{Evaluation Datasets.} We evaluate the performance of our fine-tuned models on the pushing test dataset from our system, the Object Clutter Indoor Dataset (OCID) \cite{suchi2019easylabel} and the Object Segmentation Database (OSD) \cite{richtsfeld2012segmentation}. The dataset from robot interaction is in the same domain as our collected data for fine-tuning, whereas OCID and OSD are in the different domains. The OCID dataset contains 2,390 RGB-D images, with at most 20 objects and on average 7.5 objects per image. The OSD dataset is composed of 111 RGB-D images, with up to 15 objects and an average of 3.3 objects per image.

\textbf{Evaluation Metrics.} We analyze the object segmentation performance through precision, recall, and F-measure \cite{xie2020best, xiang2021learning}. To obtain the values for these three metrics, we initially calculate the values between all pairs of predictions and ground truth objects. Subsequently, we employ the Hungarian algorithm with pairwise F-measure to match predictions with the ground truth objects.  Consequently, the precision, recall, and F-measure are determined by $P=\frac{\sum_i\left|c_i \cap g\left(c_i\right)\right|}{\sum_i\left|c_i\right|}, R=\frac{\sum_i\left|c_i \cap g\left(c_i\right)\right|}{\sum_j\left|g_j\right|}, F=\frac{2 P R}{P+R}$, where $c_i$ indicates the segmentation for the predicted object $i, g\left(c_i\right)$ is the segmentation for the corresponding ground truth object of $c_i$, and $g_j$ denotes the segmentation for the ground truth object $j$. Overlap $\mathrm{P} / \mathrm{R} / \mathrm{F}$ are the above three metrics when the intersection over union between two segmentation masks is used to determine the amount of true positives. Boundary $\mathrm{P} / \mathrm{R} / \mathrm{F}$ are also used to measure the sharpness of the predicted boundary against the ground truth boundary, where the intersection pixels of the two boundaries determines the amount of true positives. Additionally, Overlap F-measure $\geq 75 \%$ is the percentage of objects segmented with a certain
accuracy~\cite{ochs2013segmentation}.

\begin{table}[!ht]
\label{tab:uois-results-same-domain}
\caption{UOIS results on a training domain dataset. *: the model after Fine-Tuning. \textbf{MF} stands for MSMFormer \cite{lu2022mean}.}\label{same_domain_Results}
\centering
\scalebox{1.0}{
\begin{tabular}{|l|lllllll|}
\hline
\multirow{3}{*}{Method}       & \multicolumn{7}{c|}{Same Domain Dataset (107 images)}                                                                                                                                                 \\ \cline{2-8} 
                              & \multicolumn{3}{c|}{Overlap}                                                       & \multicolumn{3}{c|}{Boundary}                                                      &               \\
                              &                        \multicolumn{1}{c}{P} & \multicolumn{1}{c}{R} & \multicolumn{1}{c|}{F}             & \multicolumn{1}{c}{P} & \multicolumn{1}{c}{R} & \multicolumn{1}{c|}{F}             & \%75          \\ \hline
\multicolumn{8}{|c|}{RGB Input with ResNet-50 backbone} \\ \hline
                              
MF \cite{lu2022mean}       & 81.7                  & 81.7                  & \multicolumn{1}{l|}{81.6}          & 75.7                  & 73.1                  & \multicolumn{1}{l|}{73.7}          & 66.2          \\
MF*               & \textbf{90.6}         & \textbf{92.7}         & \multicolumn{1}{l|}{\textbf{91.6}} & \textbf{87.3}         & \textbf{88.6}         & \multicolumn{1}{l|}{\textbf{87.6}} & \textbf{90.7} \\ \hline 
MF+Zoom-in                            & 75.9                  & 81.0                  & \multicolumn{1}{l|}{78.1}          & 68.0                  & 63.7                  & \multicolumn{1}{l|}{65.1}          & 61.6          \\
MF+Zoom-in*                          & 90.1                  & 89.6                  & \multicolumn{1}{l|}{89.7}          & 88.0                  & 84.4                  & \multicolumn{1}{l|}{85.5}          & 83.5          \\
MF*+Zoom-in                            & 83.2                  & 90.9                  & \multicolumn{1}{l|}{86.7}          & 74.4                  & 78.2                  & \multicolumn{1}{l|}{75.8}          & 85.5          \\
MF*+Zoom-in*                            & \textbf{91.0}         & \textbf{93.3}         & \multicolumn{1}{l|}{\textbf{92.1}} & \textbf{89.7}         & \textbf{89.6}         & \multicolumn{1}{l|}{\textbf{89.3}} & \textbf{92.2} \\ \hline

\multicolumn{8}{|c|}{RGB-D Input with ResNet-34 backbone} \\ \hline

MF \cite{lu2022mean}          & 85.8                  & 88.9                  & \multicolumn{1}{l|}{87.2}          & 81.7                  & 78.7                  & \multicolumn{1}{l|}{79.9}          & 75.1          \\
MF*                            & \textbf{90.9}         & \textbf{91.9}         & \multicolumn{1}{l|}{\textbf{91.3}} & \textbf{86.5}         & \textbf{85.9}         & \multicolumn{1}{l|}{\textbf{85.9}} & \textbf{84.8} \\ \hline 
MF+Zoom-in                           & 88.9                  & 89.8                  & \multicolumn{1}{l|}{89.3}          & 86.6                  & 84.4                  & \multicolumn{1}{l|}{85.3}          & 80.7          \\
MF+Zoom-in*                            & 90.7                  & 90.2                  & \multicolumn{1}{l|}{90.4}          & 86.0                  & 85.9                  & \multicolumn{1}{l|}{85.6}          & 84.3          \\
MF*+Zoom-in                            & 91.0                  & \textbf{91.9}         & \multicolumn{1}{l|}{91.3}          & \textbf{89.6}         & 87.2                  & \multicolumn{1}{l|}{88.2}          & 87.0          \\
MF*+Zoom-in*                            & \textbf{92.5}         & \textbf{91.9}         & \multicolumn{1}{l|}{\textbf{92.1}} & 89.3                  & \textbf{87.8}         & \multicolumn{1}{l|}{\textbf{88.3}} & \textbf{88.0} \\  \hline
\end{tabular}
}
%\vspace{-2mm}
\end{table}

\subsection{Ablation Studies on the Fine-tuning Strategies}

%, while the pushing training data only has 321. 

We first investigate how to fine-tune the pre-trained segmentation networks with our collected real-world data. Regarding the training data for fine-tuning, we have two types of data: the 321 real-world images obtained via robot pushing and the synthetic images from the Tabletop Object Dataset~\cite{xie2020best}. The synthetic dataset consists of 280,000 RGB-D images which is used for training most unseen object instance segmentation networks~\cite{xie2020best,xiang2021learning,lu2022mean}. In this work, we use the MSMformer model~\cite{lu2022mean} trained on the Tabletop Object Dataset for fine-tuning, since it achieves very competitive performance and is end-to-end trainable. MSMformer consists of two stages in segmenting objects, where the first stage segments the whole input image while the second stage performs zoom-in refinement for each segment from the first stage.

We have two choices on using these data for fine-tuning: i) using the real-world images only, ii) using both real-world images and synthetic images. On the other hand, we have two choices on how to fine-tune the backbone network in MSMFormer: i) fixing the backbone during fine-tuning, ii) fine-tuning the backbone. We conduct ablation studies on the four combinations and present the results on the OCID and the OSD datasets in Table~\ref{backbones}.

We fine-tune the models for 6 epochs as the training loss converges quickly, where each epoch loops over the 321 real-world images once. We employ the AdamW optimizer~\cite{loshchilov2017decoupled} with the learning rate 1e-5. We set the batch size as 4. When using the mixture dataset for fine-tuning, for the first-stage model of MSMFormer, we randomly select 2 samples from the synthetic dataset and 2 samples from the real-world pushing dataset for each batch. For the second-stage model (zoom-in model), each batch has 3 random samples from the synthetic dataset and 1 pushing sample since the parameters are more sensitive to the pushing data for the second-stage.

% \yu{add some fine-tuning details here such as how many epochs to train, what the learning rate.}\lu{added above}

% The original models, MSMFormer and its zoom-in version \cite{lu2022mean}, have been trained only on synthetic data. One direct fine-tuning strategy is to train the pretrained models on pushing data. However, 

Table~\ref{backbones} shows that the performance of MSMFormer fine-tuned only using the small number of real-world pushing data is worse on the OCID dataset. This is due to overfitting to these real data. Using both the synthetic data and the real-world data for fine-tuning improves performance on both datasets. Using the mixture dataset is motivated by continual learning approaches such as \cite{aljundi2018memory,aljundi2019task} which maintains a buffer of previously seen data. In our case, we can consider the synthetic dataset to be a data buffer. Table \ref{backbones} also reveals that using learnable backbones achieves better performance than fixed backbones due to more flexibility in learning. According to these results, our fine-tuning strategy is to train the pretrained MSMFormer with mixture data and learnable backbones. We use this fine-tuning strategy in the following experiments.

% is potentially prone to overfitting. \lu{I am not sure of this part. help check} \cite{aljundi2019task} proposes that the mixture data of synthetic and pushing data can alleviate this problem since synthetic data works as a small buffer. 

% At the same time, we examine the influence of fixing the backbone and only fine-tuning the decoders on pushing data. 

\begin{table*}[!ht]
\caption{UOIS results on the OCID and OSD datasets. * indicates a model after fine-tuning. \#: \cite{back2022unseen} used different training set.}\label{all_results}
\centering
\scalebox{0.95}{
\begin{tabular}{|l|l|lllllll|lllllll|}
\hline
\multirow{3}{*}{Method}    & \multirow{3}{*}{Input} & \multicolumn{7}{c|}{OCID (2390 images)}                                                                                                                                                 & \multicolumn{7}{c|}{OSD (111 images)}                                                                                                                                                   \\ \cline{3-16} 
                           &                        & \multicolumn{3}{c|}{Overlap}                                                       & \multicolumn{3}{c|}{Boundary}                                                      &               & \multicolumn{3}{c|}{Overlap}                                                       & \multicolumn{3}{c|}{Boundary}                                                      &               \\
                           &                        & \multicolumn{1}{c}{P} & \multicolumn{1}{c}{R} & \multicolumn{1}{c|}{F}             & \multicolumn{1}{c}{P} & \multicolumn{1}{c}{R} & \multicolumn{1}{c|}{F}             & \%75          & \multicolumn{1}{c}{P} & \multicolumn{1}{c}{R} & \multicolumn{1}{c|}{F}             & \multicolumn{1}{c}{P} & \multicolumn{1}{c}{R} & \multicolumn{1}{c|}{F}             & \%75          \\ \hline
MRCNN \cite{he2017mask}                     & \multirow{10}{*}{RGB}   & 77.6                  & 67.0                  & \multicolumn{1}{l|}{67.2}          & 65.5                  & 53.9                  & \multicolumn{1}{l|}{54.6}          & 55.8          & 64.2                  & 61.3                  & \multicolumn{1}{l|}{62.5}          & 50.2                  & 40.2                  & \multicolumn{1}{l|}{44.0}          & 31.9          \\
UCN \cite{xiang2021learning}                       &                         & 54.8                  & 76.0                  & \multicolumn{1}{l|}{59.4}          & 34.5                  & 45.0                  & \multicolumn{1}{l|}{36.5}          & 48.0          & 57.2                  & \textbf{73.8}         & \multicolumn{1}{l|}{63.3}          & 34.7                  & 50.0                  & \multicolumn{1}{l|}{39.1}          & 52.5          \\
UCN+Zoom-in \cite{xiang2021learning}             &                         & 59.1                  & 74.0                  & \multicolumn{1}{l|}{61.1}          & 40.8                  & 55.0                  & \multicolumn{1}{l|}{43.8}          & 58.2          & 59.1                  & 71.7                  & \multicolumn{1}{l|}{63.8}          & 34.3                  & 53.3                  & \multicolumn{1}{l|}{39.5}          & 52.6          \\
Mask2Former~\cite{cheng2022masked}           &                         & 67.2                  & 73.1                  & \multicolumn{1}{l|}{67.1}          & 55.9                  & 58.1                  & \multicolumn{1}{l|}{54.5}          & 54.3          & 60.6                  & 60.2                  & \multicolumn{1}{l|}{59.5}          & 48.2                  & 41.7                  & \multicolumn{1}{l|}{43.3}          & 32.4          \\
MF~\cite{lu2022mean}            &                         & 72.9                  & 68.3                  & \multicolumn{1}{l|}{67.7}          & 60.5                  & 56.3                  & \multicolumn{1}{l|}{55.8}          & 52.9          & 63.4                  & 64.7                  & \multicolumn{1}{l|}{63.6}          & 48.6                  & 47.4                  & \multicolumn{1}{l|}{47.0}          & 40.2          \\
MF*    &                         & 80.3                  & 81.0                  & \multicolumn{1}{l|}{78.5}          & 67.9                  & 70.8                  & \multicolumn{1}{l|}{67.4}          & 73.1          & 63.5                  & 71.7                  & \multicolumn{1}{l|}{66.8}          & 44.8                  & 56.2                  & \multicolumn{1}{l|}{48.7}          & 49.8          \\
MF+Zoom-in~\cite{lu2022mean}   &                         & 73.9                  & 67.1                  & \multicolumn{1}{l|}{66.3}          & 64.6                  & 52.9                  & \multicolumn{1}{l|}{54.8}          & 52.8          & 63.9                  & 63.7                  & \multicolumn{1}{l|}{62.7}          & 51.6                  & 45.3                  & \multicolumn{1}{l|}{47.0}          & 41.1          \\
MF+Zoom-in*  &                         & 76.6                  & 71.1                  & \multicolumn{1}{l|}{70.4}          & 68.1                  & 57.8                  & \multicolumn{1}{l|}{59.8}          & 58.7          & \textbf{69.8}         & 69.1                  & \multicolumn{1}{l|}{\textbf{69.0}} & \textbf{54.8}         & 52.2                  & \multicolumn{1}{l|}{\textbf{52.5}} & 50.7          \\
MF*+Zoom-in   &                         & 79.5                  & 77.1                  & \multicolumn{1}{l|}{75.5}          & 68.9                  & 64.8                  & \multicolumn{1}{l|}{64.6}          & 68.3          & 63.4                  & 71.5                  & \multicolumn{1}{l|}{66.6}          & 47.0                  & 55.0                  & \multicolumn{1}{l|}{49.4}          & 52.0          \\
MF*+Zoom-in* &                         & \textbf{83.5}         & \textbf{82.1}         & \multicolumn{1}{l|}{\textbf{80.6}} & \textbf{74.1}         & \textbf{72.5}         & \multicolumn{1}{l|}{\textbf{71.7}} & \textbf{77.9} & 66.2                  & 72.3                  & \multicolumn{1}{l|}{68.4}          & 50.3                  & \textbf{56.8}         & \multicolumn{1}{l|}{52.0}          & \textbf{56.0} \\ \hline
MRCNN \cite{he2017mask}                     & \multirow{3}{*}{Depth}  & 85.3                  & 85.6                  & \multicolumn{1}{l|}{84.7}          & 83.2                  & 76.6                  & \multicolumn{1}{l|}{78.8}          & 72.7          & 77.8                  & 85.1                  & \multicolumn{1}{l|}{80.6}          & 52.5                  & 57.9                  & \multicolumn{1}{l|}{54.6}                               & 77.6          \\
UOIS-Net-2D \cite{xie2020best}       &                         & 88.3                  & 78.9                  & \multicolumn{1}{l|}{81.7}          & 82.0                  & 65.9                  & \multicolumn{1}{l|}{71.4}          & 69.1          & 80.7                  & 80.5                  & \multicolumn{1}{l|}{79.9}          & 66.0                  & 67.1                  & \multicolumn{1}{l|}{65.6}                               & 71.9          \\
UOIS-Net-3D \cite{xie2021unseen}       &                         & 86.5                  & 86.6                  & \multicolumn{1}{l|}{86.4}          & 80.0                  & 73.4                  & \multicolumn{1}{l|}{76.2}          & 77.2          & 85.7                  & 82.5                  & \multicolumn{1}{l|}{83.3}          & 75.7                  & 68.9                  & \multicolumn{1}{l|}{71.2}                              & 73.8          \\ \cline{2-2}
MRCNN \cite{he2017mask}                     & \multirow{11}{*}{RGB-D} & 79.6                  & 76.7                  & \multicolumn{1}{l|}{76.6}          & 68.7                  & 63.7                  & \multicolumn{1}{l|}{64.3}          & 62.9          & 66.4                  & 64.8                  & \multicolumn{1}{l|}{65.5}          & 53.7                  & 43.8                  & \multicolumn{1}{l|}{47.5}                               & 37.1          \\
UCN \cite{xiang2021learning}                          &                         & 86.0                  & \textbf{92.3}         & \multicolumn{1}{l|}{88.5}          & 80.4                  & 78.3                  & \multicolumn{1}{l|}{78.8}          & 82.2          & 84.3                  & \textbf{88.3}         & \multicolumn{1}{l|}{86.2}          & 67.5                  & 67.5                  & \multicolumn{1}{l|}{67.1}                               & 79.3          \\
UCN+Zoom-in \cite{xiang2021learning}             &                         & 91.6                  & 92.5                  & \multicolumn{1}{l|}{91.6}          & 86.5                  & \textbf{87.1}         & \multicolumn{1}{l|}{86.1}          & \textbf{89.3} & 87.4                  & 87.4                  & \multicolumn{1}{l|}{\textbf{87.4}} & 69.1                  & 70.8                  & \multicolumn{1}{l|}{69.4}                               & 83.2          \\
UOAIS-Net \cite{back2022unseen} \#                &                         & 70.7                  & 86.7                  & \multicolumn{1}{l|}{71.9}          & 68.2                  & 78.5                  & \multicolumn{1}{l|}{68.8}          & 78.7          & 85.3                  & 85.4                  & \multicolumn{1}{l|}{85.2}          & \textbf{72.7}         & \textbf{74.3}         & \multicolumn{1}{l|}{\textbf{73.1}}                      & 79.1          \\
Mask2Former \cite{cheng2022masked}               &                         & 78.6                  & 82.8                  & \multicolumn{1}{l|}{79.5}          & 69.3                  & 76.2                  & \multicolumn{1}{l|}{71.1}          & 69.3          & 75.6                  & 79.2                  & \multicolumn{1}{l|}{77.3}          & 54.1                  & 64.0                  &  \multicolumn{1}{l|}{58.0}                              & 65.2          \\
MF~\cite{lu2022mean}                 &                         & 88.4                  & 90.2                  & \multicolumn{1}{l|}{88.5}          & 84.7                  & 83.1                  & \multicolumn{1}{l|}{83.0}          & 80.3          & 79.5                  & 86.4                  & \multicolumn{1}{l|}{82.8}          & 53.5                  & 71.0                  & \multicolumn{1}{l|}{60.6}                              & 79.4          \\
MF*         &                         & 91.2                  & 90.1                  & \multicolumn{1}{l|}{90.1}          & 87.2                  & 85.5                  & \multicolumn{1}{l|}{85.7}          & 83.9          & 85.1                  & 84.4                  & \multicolumn{1}{l|}{84.6}          & 67.8                  & 71.4                  & \multicolumn{1}{l|}{69.0}                               & 76.2          \\
MF+Zoom-in \cite{lu2022mean}       &                         & 92.5                  & 91.0                  & \multicolumn{1}{l|}{91.5}          & 89.4                  & 85.9                  & \multicolumn{1}{l|}{87.3}          & 86.0          & 87.1                  & 86.1                  & \multicolumn{1}{l|}{86.4}          & 69.0                  & 68.6                  & \multicolumn{1}{l|}{68.4}                               & \textbf{80.4} \\
MF+Zoom-in*       &                         & 91.3                  & 91.3                  & \multicolumn{1}{l|}{91.0}          & 87.7                  & 84.4                  & \multicolumn{1}{l|}{85.6}          & 86.2          & 86.6                  & 82.9                  & \multicolumn{1}{l|}{84.4}          & 68.9                  & 69.6                  & \multicolumn{1}{l|}{68.6}                               & 76.3          \\
MF*+Zoom-in      &                         & \textbf{92.7}         & 91.6                  & \multicolumn{1}{l|}{\textbf{91.9}} & \textbf{89.9}         & 86.5                  & \multicolumn{1}{l|}{\textbf{87.8}} & 87.1          & 85.5                  & 84.6                  & \multicolumn{1}{l|}{84.8}          & 68.3                  & 65.9                  & \multicolumn{1}{l|}{66.7}                               & 75.5          \\
MF*+Zoom-in*      &                         & 91.5                  & 91.8                  & \multicolumn{1}{l|}{91.3}          & 88.2                  & 84.8                  & \multicolumn{1}{l|}{86.1}          & 87.1          & \textbf{87.6}         & 82.6                  & \multicolumn{1}{l|}{84.4}          & 70.8                  & 68.5                  & \multicolumn{1}{l|}{68.9}                              & 75.8          \\ \hline
\end{tabular}
}
\end{table*}

\begin{figure*}
\begin{center}
\includegraphics[width=0.95\linewidth]{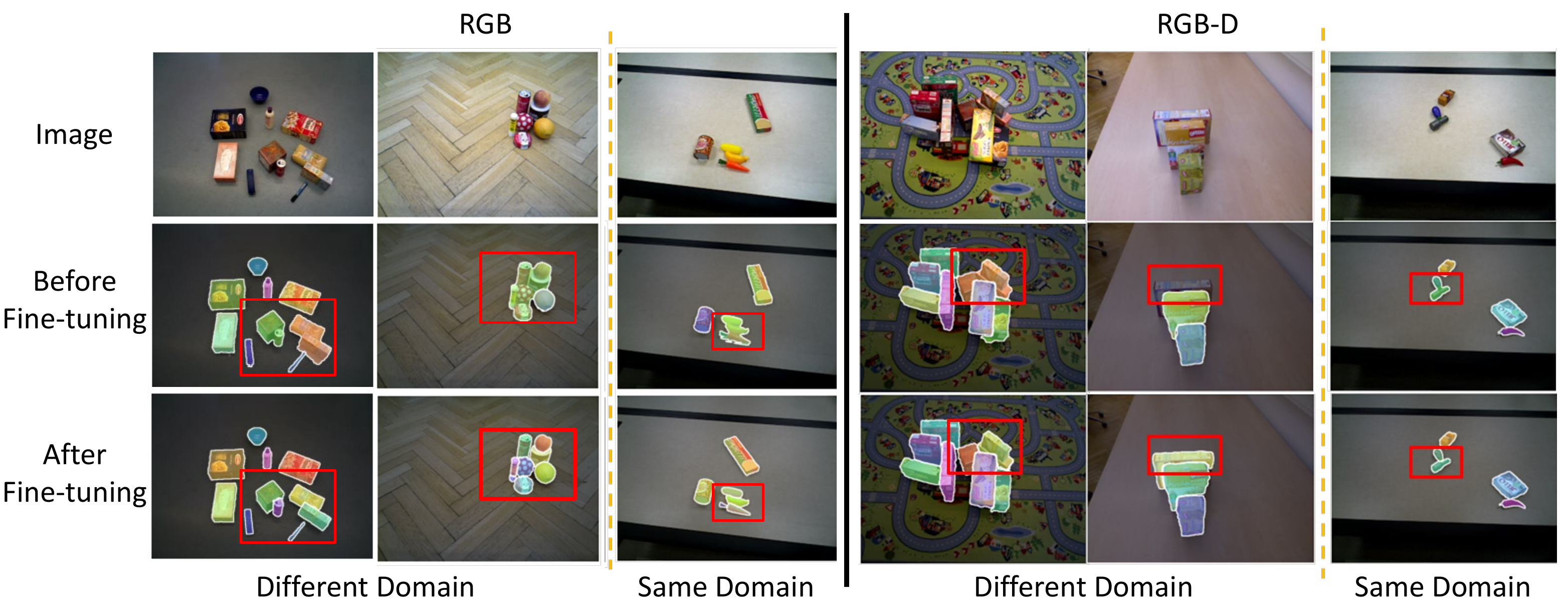}
\caption{Illustration of the effect of fine-tuning of the MSMFormer. The fine-tuning of the model allows it to distinguish objects that are stacked or adjacent to each other, where the original model cannot separate these objects.}
\label{fig:diff}
\end{center}
\vspace{-6mm}
\end{figure*}

\subsection{Ablation Studies on the Number of Fine-tuning Images}

Our collected pushing training set has 15 scenes in total. We investigate the correlation between the number of images and the performance of the fine-tuned model. We partition the training set according to scenes and gradually add more scenes to the fine-tuning dataset. Table~\ref{scenes} shows the performance of the MSMFormer models fine-tuned with datasets in different sizes. We can see that, the performance on the OCID and OSD datasets continually improves as the amount of scenes expands. After 12 scenes, the model performance begins to saturate. According to this experiment, a small number of real-world images for fine-tuning is sufficient, which avoids collecting a large number of images in the real world for fine-tuning. We use all the 15 scenes with 321 images for fine-tuning in the following experiments.

% plateau out. The performance on OCID is potentially improving with more scenarios.

% We are also curious about how many samples are necessary to guarantee a high-quality model. Table \ref{scenes} illustrates that 

\subsection{Object Instance Segmentation in the Same Domain}

Table~\ref{same_domain_Results} presents the evaluation results on the 107 real-world test images of the models before and after fine-tuning. Since the pushing test dataset has the same settings as the fine-tuning dataset, we view the pushing test dataset as in same domain.
It is clear that the fine-tuned models significantly improve the segmentation accuracy in the same domain. Imagine a robot entering a new domain, it can utilize our system to collect a few images to improve object segmentation in this new domain. We experiment fine-tuning both the RGB version and the RGB-D version of MSMFormer. In addition, we investigate the effect of fine-tuning on each stage of the segmentation network. ``Zoom-in'' in Table~\ref{same_domain_Results} indicates the second-stage network. From the table, we can see that fine-tuning consistently improves the performance over the original models. The best performance is achieved by fine-tuning both stages of MSMFormer.

Generally, RGB-D models tend to surpass RGB models due to the additional depth input. However, we can observe that the fine-tuned two-stage RGB model (RGB with zoom-in) achieves the same Overlap F-measure and a higher Boundary F-measure compared to the fine-tuned two-stage RGB-D model. This result indicates that it is possible to segment unseen objects with RGB images only as long as we can obtain RGB training images with ground truth labels. Our system provides a solution by utilizing robot interaction for data collection. It is worth noting that using RGB images only is valuable since certain objects such as transparent objects or metal objects cannot be captured well by depth images.

% a Boundary F-measure of 89.3, higher than the Boundary F-measure of 88.3 from the RGB-D model. This implies that the RGB model can be useful for UOIS under certain conditions, especially when it is challenging to acquire depth images. 

% There are four distinct cases: RGB, RGB with zoom-in, RGB-D, and RGB-D with zoom-in. In all four cases, most metrics indicate a substantial increase compared to the baseline model in Table \ref{same_domain_Results}. 

\subsection{Object Instance Segmentation across Domains}
To evaluate the performance of the fine-tuned models across domains, we test them on the OCID and OSD datasets and compare the achieved results with the state-of-the-art methods in Table \ref{all_results}. From the table, we can see that the fine-tuned models improve over the state-of-the-art methods on the OCID dataset for both RGB and RGB-D input. On the OSD dataset, UOAIS-Net \cite{back2022unseen} achieves better performance for RGB-D input by utilizing photo-realistic synthetic images for training.

% The fine-tuned first-stage model performs slightly better than the original one. Especially on OSD, the RGB-D first-stage model achieve 69.0 Boundary F-measure, surpassing the result of the model with zoom-in. It suggests that it is feasible to utilize only one stage for successful predictions. 

In most cases, the fine-tuning strategy consistently improves the pre-trained models with synthetic images. However, the RGB-D fine-tuned zoom-in refinement is not as effective as the original zoom-in refinement on the OCID dataset. The primary reason for this is that the environment and objects in our pushing dataset are simpler and more restricted than those presented in the OCID dataset. The combination of the fine-tuned first-stage model and the original zoom-in part is more effective on the OCID dataset. We visualize the differences of using the original models and fine-tuned models on different datasets in Fig. \ref{fig:diff}. The fine-tuned models are able to separate adjacent objects to mitigate the under-segmentation problem in the same domain as the fine-tuning images as well as different domains in the OCID and OSD datasets.

%%%% Grasping Experiments: 4 different scenes %%%%

%%%% Grasping Experiments Figure %%%%
\begin{figure*}
\begin{center}
\includegraphics[width=0.95\linewidth]{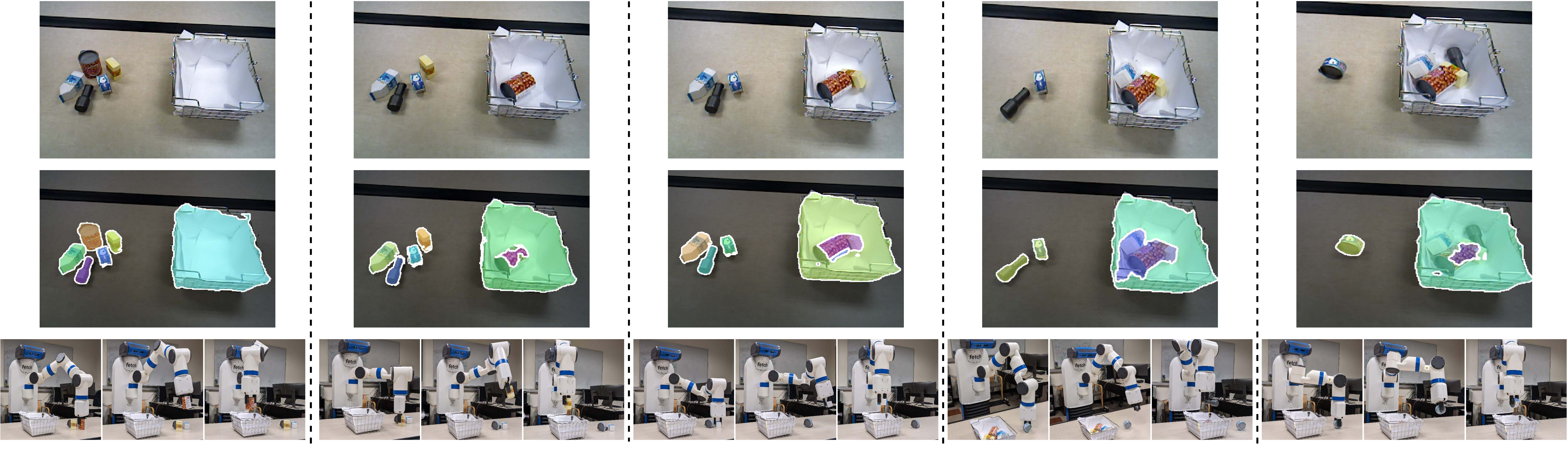}
\caption{Setup for top-down grasping with segmentation. Robot images on each column show the three stages: approach, pickup and placing in the bin.}
\label{fig:top-down-grasping-experiments}
\end{center}
\vspace{-6mm}
\end{figure*}

\begin{figure}
  \begin{minipage}{0.57\linewidth}
    \captionof{figure}{Examples of scene configurations with varying amount of clutter.\vspace{5pt}}
    \includegraphics[width=\linewidth]{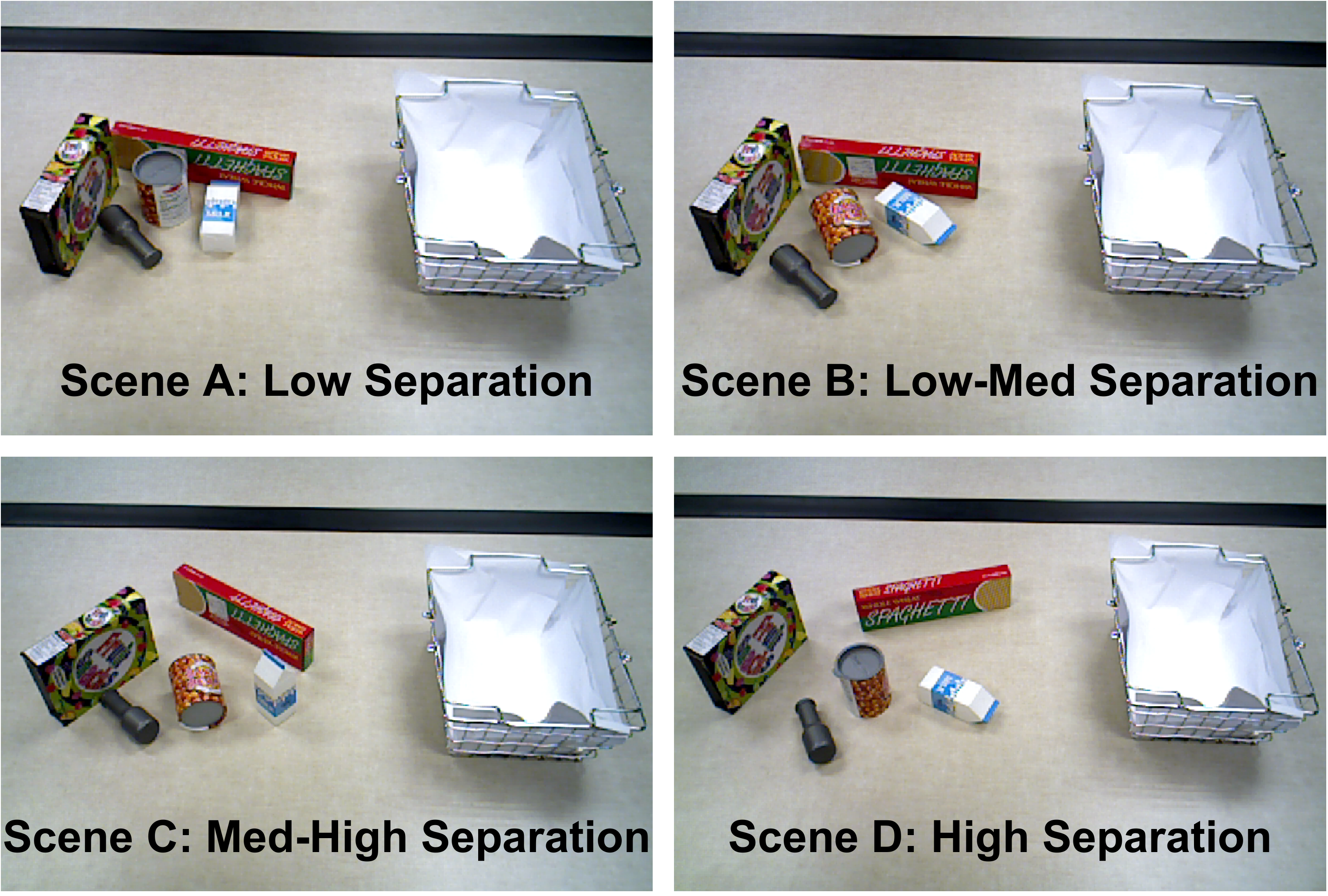}
    \label{fig:grasp-exp-4-scenes}
    \centering
  \end{minipage}
  \hfill
  \begin{minipage}{0.42\linewidth}
    \centering
    \captionof{table}{Grasp success with different scene configurations.}
    \scalebox{0.65}{
    \begin{tabular}{|l|c|c|}
        \hline
        Obj. Set       & Baseline\textsuperscript{\ref{fnote:baseline}}       & Fine-tune\textsuperscript{\ref{fnote:fine-tune}}      \\ \hline
        1-A              & 0              & 5              \\
        1-B              & 0              & 5              \\
        1-C              & 1              & 3              \\
        1-D              & 5              & 4              \\ \hline
        Total          & 6              & 17             \\ \hline
        2-A              & 4              & 4              \\
        2-B              & 1              & 3              \\
        2-C              & 1              & 5              \\
        2-D              & 5              & 3              \\ \hline
        Total          & 11             & 15             \\ \hline
        \textbf{Overall} & \textbf{17/40} & \textbf{32/40} \\ \hline
    \end{tabular}
    \label{tab:grasping-exp-results}} %end the scalebox env
  \end{minipage}
  \vspace{-4mm}
\end{figure}

\subsection{Top-Down Grasping with Object Instance Segmentation}
\label{sec:exp-top-down-grasping}

We show the usefulness of the proposed system for grasping unknown objects in a table-top setting where the objects are placed in a cluttered environment. A Fetch mobile manipulator is used for the experiments with its parallel jaw gripper for grasping, and built in RGB-D camera for perception. We compute the top-down grasp after segmenting all the objects in the scene via the procedure described in Section~\ref{sec:applications-top-down-grasping}. We formulate the experiment as a pick-and-place task where the goal is to clear the table and place all the objects in a nearby bin. One example is shown in Fig.~\ref{fig:top-down-grasping-experiments}.

The experiment is conducted with two sets of unknown objects (i.e., not seen during fine-tuning or training) with each set containing five objects. For each object set, we consider the pick-and-place task with four different initial configurations of the object placement on the table, ranging from highly cluttered to well separated as shown in Fig.~\ref{fig:grasp-exp-4-scenes}. The pick-and-place grasping trials are conducted with the baseline\footnote{\label{fnote:baseline}Baseline: MSMFormer\_R34 + Zoom-in in Table~\ref{tab:uois-results-same-domain}} and fine-tuned\footnote{\label{fnote:fine-tune}Fine-tuned: MSMFormer\_R34* + Zoom-in* in Table~\ref{tab:uois-results-same-domain}} segmentation models with RGB-D input for each configuration to bring out the relative improvement of fine-tuning on data collected using the proposed method. 

Given a configuration for object arrangement on the table, there are five pick-and-place trials associated with each of the five objects. A trial is counted as a success if a grasp of an object guided by its segmentation boundary allows for a successful pick-and-place operation, otherwise its counted as a failure.
A hard-failure occurs for a scene if the segmentation masks are incorrect in the beginning, with the 5 objects in the scene. 
Such an error is not favorable due to possibility of collision and damage of the gripper and hence the grasping is stopped in this case, and none of the objects count towards the success rate metric. It potentially occurs if the segmentation model is not able to establish clear boundaries between nearby objects which induces errors in the grasping pipeline, specifically in positioning the gripper for picking up the object. 
For example, cases 1-A and 1-B with the baseline model in Table~\ref{tab:grasping-exp-results} are hard failures due to segmentation error at the very start. Consequently, no feasible grasping motion is found for any object in the scene and hence they have no score for the respective trials. Therefore, accurate object segmentation is critical for grasping in cluttered scenes.

We obtain data for the 40 individual trials (10 objects in total, across 4 table-top configurations for each) for each of the baseline and fine-tuned models and report their number of successful actions. As seen in Table~\ref{tab:grasping-exp-results}, we see a clear improvement in the grasp success rate when using the fine-tuned model, especially in scenes with high clutter. This highlights the need for precise segmentation masks of objects in cluttered scenes as any errors in this stage likely affect downstream applications like grasping.   
Additional details and qualitative results will be provided in the supplementary material.

\section{Conclusion and Future Work} 
\label{sec:conclusion}

We introduced a robotic system for self-supervised unseen object instance segmentation. Our system leverages robot pushing to interact with objects and collect images before and after each pushing action. In order to generate segmentation masks of objects in the collected images, the system allows the robot to push objects until a sequence of images is collected, then an optical flow based multi-object tracking algorithm and a video object segmentation method are combined to segment object instances in the collected images automatically. Using a sequence of images from robot pushing enables the system to segment all the objects in the sequence including objects that are very close to each other. To the best of our knowledge, this is a first system that leverages long-term robot interaction for object segmentation.

We verify the usefulness of the system by using the collected images to fine-tune object segmentation networks. Our experiments show that the fine-tuned networks achieve better segmentation accuracy both in the same domain and in different domains. We also demonstrate that improving object segmentation with fine-tuning benefit top-down robot grasping in a pick-and-place task, where accurate object segmentation can be used to plan grasps in cluttered scenes.

For future work, we plan to extend the system beyond tabletop scenarios such as segmenting objects inside bins or cabinets. Robot interaction in these environments requires motion planning to account for the constraints from the environments. Robot pushing may not be sufficient in these environments. We plan to investigate different interaction actions such as grasping and scooping for data collection.

\section*{Acknowledgments}

This work was supported in part by the DARPA Perceptually-enabled Task Guidance (PTG) Program under contract number HR00112220005. Kaiyu Hang is supported by NSF CMMI-2133110.

%% Use plainnat to work nicely with natbib. 

\bibliographystyle{plainnat}
\bibliography{references}

\end{document}